%% file: paper-v1.tex
\documentclass[journal]{IEEEtran}
\usepackage[super]{nth}
\usepackage{hyperref}
\usepackage{cite}
\usepackage[pdftex]{graphicx}
\usepackage{subcaption}
\usepackage{float}
\usepackage{balance}
\usepackage[ruled,vlined]{algorithm2e}

\usepackage{xspace}

\usepackage{amsmath}
\usepackage{amssymb}
\usepackage{tabularx}
\usepackage{makecell}
\newcolumntype{C}[1]{>{\centering\arraybackslash}p{#1}}
\newcolumntype{Y}{>{\centering\arraybackslash}X}

\usepackage{booktabs}
\usepackage{multirow}
\usepackage[table]{xcolor}
\usepackage{pifont}

\usepackage{comment}
\usepackage{mathtools}

\usepackage{svg}
\graphicspath{ {./images/} }

\usepackage{acronym}
    \acrodef{ML}{Machine Learning}
    \acrodef{CL}{Continual Learning}
    \acrodef{FL}{Federated Learning}
    \acrodef{FCL}{Federated Continual Learning}
    \acrodef{PFL}{Personalized Federated Learning}
    \acrodef{non-IID}{non-independent and identically distributed}
    \acrodef{MoE}{Mixture-of-Experts}

\newcommand{\method}{\texttt{DriftGuard}\xspace}

\newcommand{\effimpt}{{2.3$\times$}\xspace} 

\newcommand{\costred}{{22\%--57\%}\xspace}

\newcommand{\costredc}{{22\%-83\%}\xspace}
\newcommand{\costredT}{{83\%}\xspace}

\begin{document}

\title{\method: Mitigating Asynchronous Data Drift in Federated Learning}

\author{
    \IEEEauthorblockN{
    Yizhou Han\IEEEauthorrefmark{1},
    Di Wu\IEEEauthorrefmark{1},
    and Blesson Varghese\IEEEauthorrefmark{1}\\
    }
    \IEEEauthorblockA{
    \IEEEauthorrefmark{1}\textit{School of Computer Science, University of St Andrews, UK}
    }
    \thanks{Corresponding author: Di Wu (E-mail: dw217@st-andrews.ac.uk).}
    \thanks{Yizhou Han, Di Wu, and Blesson Varghese are with the School of Computer Science, University of St Andrews, UK.}
    \thanks{
	    Copyright (c) 20xx IEEE. Personal use of this material is permitted. However, permission to use this material for any other purposes must be obtained from the IEEE by sending a request to pubs-permissions@ieee.org.
	    }
}

\maketitle
\thispagestyle{plain}
\pagestyle{plain}

\begin{abstract}
\input{sections/abstract}
\end{abstract}


\IEEEpeerreviewmaketitle

\section{Introduction}
\label{sec:introduction}
\input{sections/introduction}

\section{Background and Motivation}
\label{sec:background}
\input{sections/background}

\section{Problem Formulation}
\label{sec:model}
\input{sections/model}

\section{\method}
\label{sec:method}
\input{sections/method}

\section{Evaluation}
\label{sec:evaluation}
\input{sections/evaluation}

\section{Related Work}
\label{sec:relatedwork}
\input{sections/relatedwork}

\section{Conclusions}
\label{sec:conclusion}

\input{sections/conclusion}

\section*{Acknowledgment}
\label{sec:acknowledgment}
\input{sections/acknowledgment}

\balance
\bibliographystyle{IEEEtran}
\bibliography{references}

\input{sections/authorsbiography}





\end{document}

%% file: sections/abstract.tex
In real-world Federated Learning (FL) deployments, data distributions on devices that participate in training evolve over time. This leads to asynchronous data drift, where different devices shift at different times and toward different distributions. Mitigating such drift is challenging: frequent retraining incurs high computational cost on resource-constrained devices, while infrequent retraining degrades performance on drifting devices.
We propose \method, a federated continual learning framework that efficiently adapts to asynchronous data drift. \method adopts a Mixture-of-Experts (MoE) inspired architecture that separates shared parameters, which capture globally transferable knowledge, from local parameters that adapt to group-specific distributions. This design enables two complementary retraining strategies: (i) global retraining, which updates the shared parameters when system-wide drift is identified, and (ii) group retraining, which selectively updates local parameters for clusters of devices identified via MoE gating patterns, without sharing raw data.
Experiments across multiple datasets and models show that \method matches or exceeds state-of-the-art accuracy while reducing total retraining cost by up to \costredT. As a result, it achieves the highest accuracy per unit retraining cost, improving over the strongest baseline by up to \effimpt.
\method is available for download from \url{https://github.com/blessonvar/DriftGuard}.

%% file: sections/introduction.tex
The proliferation of Internet of Things (IoT) devices, such as smart cameras and wearable sensors, has led to the continuous generation of data at the network edge~\cite{zhang2021sv-iot, pan2017iotj_future}.
\ac{FL} has emerged as a paradigm that trains machine learning models locally on devices without requiring the exchange of raw data to extract knowledge from edge-generated data~\cite{zhang2021sv-fl, imteaj2021iotj_fl}.
However, conventional \ac{FL} assumes stationary data distributions on devices, which is often not the case in practice as data distributions in the real-world evolve over time with newly generated samples; this is referred to as data drift~\cite{yang2024sv-fcl-fusion,gama2014sv-conceptdrift}.

To mitigate data drift, \ac{CL} retrains models on newly arriving data samples to adapt to evolving data distributions~\cite{lu2018conceptdrift}. \ac{FL} systems incorporate \ac{CL} by periodically or conditionally retraining when the global model accuracy degrades to address data drift on devices.
However, FL retraining is challenging, as it requires multiple rounds of local training on resource-constrained devices, leading to significant computational overheads~\cite{nguyen2021fl_for_iot}.

Moreover, in real-world \ac{FL} systems, data drift occurs asynchronously across the devices - each device has a different drift rate and drifts towards different distributions. This is referred to as \textit{asynchronous data drift}, which is fundamentally different from \textit{synchronous data drift} that is typically explored in traditional \ac{CL}~\cite{jothimurugesan2023feddrift}. An example of asynchronous data drift is a smart-city traffic monitoring application in which only a subset of cameras may experience data drift due to localized changes in weather conditions or illumination~\cite{snyder2019streets}. 

Consequently, mitigating data drift in \ac{FL} systems is non-trivial.
Since FL retraining incurs significant system overhead (e.g., computational cost), minimizing the frequency of retraining is desirable.
However, identifying an efficient retraining configuration (i.e., when to retrain and which devices should participate in retraining) is challenging under asynchronous data drift, as some devices may experience rapid distribution shifts while others remain relatively stable with limited distribution change.
This leads to a fundamental trade-off: frequent retraining strategies incur substantial system overheads, whereas conservative retraining may result in drifting devices underperforming for extended periods.

Existing approaches to address the above fall into two categories: traditional \ac{FCL} and clustering-based \ac{FCL}.
Traditional \ac{FCL} methods~\cite{mcmahan2017FedAvg, li2020FedProx, reddi2021FedOpt} rely on frequent global retraining on all devices. This results in large system overheads when data drift occurs asynchronously across devices.
Clustering-based approaches~\cite{jothimurugesan2023feddrift, peng2025feddaa} restrict retraining to subsets of devices with similar data drift patterns, thereby reducing the number of participating devices and lowering the overall retraining cost.
However, by treating each cluster as an independent and mutually exclusive unit, these methods limit the sharing of globally transferable knowledge, resulting in suboptimal accuracy.

In this article, we propose \method, a novel framework that mitigates asynchronous data drift in \ac{FL} systems while significantly reducing retraining cost and maintaining model accuracy.
\method adopts a \ac{MoE} inspired architecture deployed on devices, which comprises shared parameters, which captures globally transferable knowledge, and local parameters, which adapt to group-specific data distributions. By leveraging this architecture, \method decouples FL retraining into two complementary strategies:
(i) \textit{Global retraining}, which updates the shared parameters only when global drift is detected; and
(ii) \textit{Group retraining}, which selectively updates local parameters for a group of devices that experience similar data drift.
To identify device groups with similar data distributions, \method leverages MoE gating outputs to cluster devices without requiring them to exchange raw data.
By triggering global retraining only when globally transferable knowledge becomes outdated and performing group retraining only for small device groups, \method maintains high model accuracy while reducing retraining overhead compared to traditional FCL.

Experiments across three datasets with four models show that \method achieves the highest or comparable accuracy to the strongest baselines while reducing total retraining cost by up to \costredT. This leads to the highest accuracy per retraining cost ($\mathcal{E}$), up to \effimpt higher than the strongest baseline. On a real-world IoT prototype, \method consistently delivers the highest accuracy across all experimental settings while reducing retraining time by up to 20\%, resulting in the highest $\mathcal{E}$, up to 1.2$\times$ higher than the strongest baseline.

The key contributions of this article are as follows:
\begin{enumerate}
    \item \method, a novel \ac{FCL} framework that leverages an \ac{MoE} inspired architecture to separate globally transferable knowledge from group-specific adaptations, enabling efficient retraining under asynchronous data drift.
    \item A device grouping method based on \ac{MoE} gating outputs, allowing devices with similar data distributions to be clustered without sharing raw data.
    \item A two-level retraining mechanism that selectively performs \textit{global retraining} and \textit{group retraining}, striking a balance between model accuracy and system overhead.
\end{enumerate}

The rest of this article is organized as follows. 
Section~\ref{sec:background} introduces data drift and continual learning, and identifies the challenge of asynchronous data drift in federated learning. 
Section~\ref{sec:model} formulates the problem and defines the optimization objective. 
Section~\ref{sec:method} presents the design of \method. 
Section~\ref{sec:evaluation} evaluates \method on across multiple datasets and models. 
Section~\ref{sec:relatedwork} discusses related work, and 
Section~\ref{sec:conclusion} concludes the article.

%% file: sections/background.tex

\subsection{Data Drift and Continual Learning}
Data drift is a shift in the input data distribution over time relative to the training distribution~\cite{gama2014sv-conceptdrift, lu2018conceptdrift}.
This is a common occurrence in real-world deployments of long-running ML systems, where data-generating processes evolve over time~\cite{Hao2025nazar}. For example, changes in user behavior or environmental conditions can shift the data distribution~\cite{gama2014sv-conceptdrift, dong2024efficiently}.
Since such drift typically occurs after model training, the trained model will no longer fit the new input data distribution well. This results in reduced accuracy on incoming data and degrades the quality of the \ac{ML} system.

To address the aforementioned issues caused by data drift, retraining on new incoming data is required to adapt the model to the evolving data distribution. This is commonly referred to as \ac{CL} in the literature.
However, retraining ML models, especially deep learning models, typically incurs substantial computational costs, making continual retraining expensive and less practical. A key challenge in \ac{CL} is therefore how to reduce training costs while adapting to gradual data distribution shifts. To this end, various approaches have been proposed, such as fine-tuning previously trained models instead of retraining from scratch~\cite{kirkpatrick2017ewc, lopez2017gem} and parameter isolation and reuse~\cite{mallya2018packnet}.
Most of the existing \ac{CL} research assumes that retraining is performed in centralized settings, where all newly arriving data are collected and stored at a single location.

\subsection{Asynchronous Data Drift in Federated Learning}
FL trains a global model on distributed devices by performing local updates on each device and aggregating these updates on a central server~\cite{mcmahan2017FedAvg, zhang2021sv-fl, wen2023sv_fl}. This offers privacy benefits compared to traditional centralized training, as raw data remain on local devices rather than being transferred to the server.

\begin{figure}[t]
  \centering
  \includegraphics[width=\linewidth]{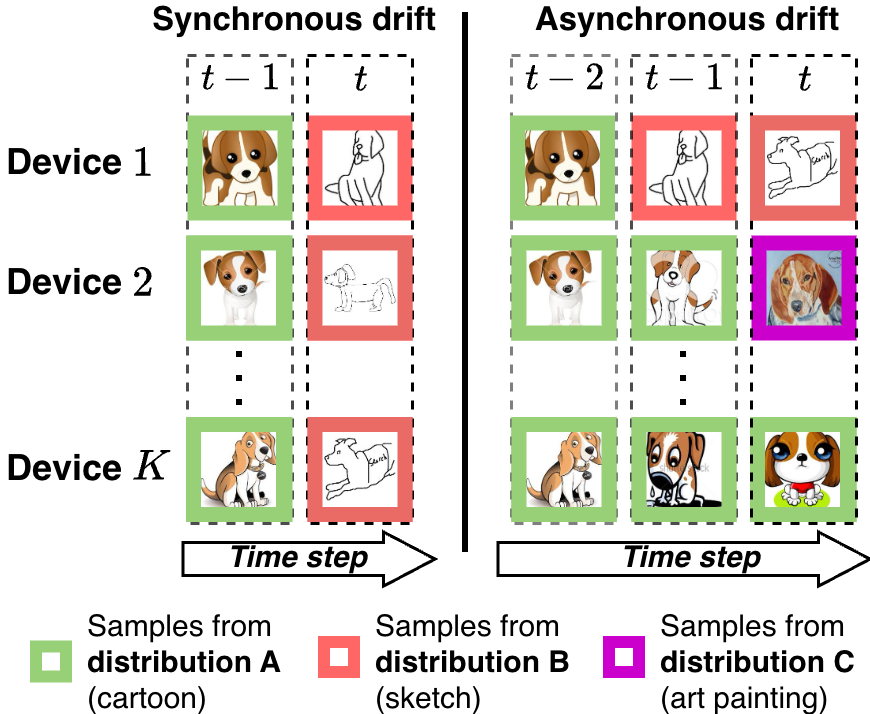}
  \caption{Illustration of synchronous and asynchronous data drift.}
  \label{fig:asynchronous}
\end{figure}

Data drift can also occur in \ac{FL} systems since the data distribution on devices may change over time. Importantly, such changes typically occur independently on devices, leading to \textbf{asynchronous data drift}. 
Figure~\ref{fig:asynchronous} illustrates this phenomenon conceptually on a set of devices $K$, using image style domains, such as cartoon, sketch, and art painting, to represent distinct data distributions.
Unlike synchronous drift where all devices shift to the same distribution simultaneously in a time step, asynchronous drift occurs when different devices drift to different distributions at different times.
In real world federated IoT systems, asynchronous data drift occurs more commonly than synchronous data drift as each device may often operate in distinct environments~\cite{jothimurugesan2023feddrift}. For example, in smart-city traffic monitoring, cameras at different intersections can experience different weather and lighting conditions over time, causing drift on different subsets of devices at different times~\cite{snyder2019streets}.

Mitigating asynchronous data drift in FL is more challenging than handling synchronous data drift in traditional CL for two reasons.
First, it is difficult to determine an appropriate retraining schedule, as data drift may occur at different rates on devices: some devices may experience rapid distribution changes, while others may exhibit little or no drift.
Second, retraining in FL is typically more costly than in centralized settings, as it requires training on a large number of devices and often involves more communication rounds than centralized retraining~\cite{imteaj2021iotj_fl}.

Existing approaches to mitigating asynchronous data drift in \ac{FL} have notable limitations.
Traditional \ac{FCL} methods trigger global retraining on all devices, incurring high overhead when only a subset experiences drift. Clustering-based methods reduce cost by grouping devices but treat groups independently, failing to leverage globally transferable knowledge to achieve higher accuracy. 
These limitations motivate designing new strategies for configuring retraining that decouple globally transferable knowledge from group-specific adaptations, enabling selective retraining to improve global model accuracy while reducing retraining costs.

%% file: sections/model.tex
In this section, we first formulate asynchronous data drift in FL systems and present a unified retraining framework that can be used to incorporate different retraining strategies for FL. We analyze the retraining costs in FL and define the optimization objective within this framework. This provides a unified basis for characterizing and comparing different retraining strategies in terms of model accuracy and computational costs.

\subsection{Asynchronous Data Drift Model}
An FL system consists of a central server and $K$ devices. 
The set of devices is denoted by $\mathcal{C}=\{1,2,\dots,K\}$. 
We assume that each device $c\in\mathcal{C}$ continuously generates new local data over time, causing its data distribution to evolve at each time step $t$. 
Specifically, device $c$ generates a new local dataset $\mathcal{D}_c^{t}$ based on an underlying distribution $P_c^{t}$. We use \textit{local} to refer to on-device quantities, such as data, model, and accuracy.

Under \textit{asynchronous data drift}, the underlying data distribution of newly generated data on device $c$ may vary across time steps, i.e., 
$\exists\, i \neq j \text{ such that } P_c^{i} \neq P_c^{j}$. 
In addition, the data distributions may also vary on different devices at the same time step $t$, i.e., 
$\exists\, c_i \neq c_j \text{ such that } P_{c_i}^{t} \neq P_{c_j}^{t}$.
As a result, data drift may occur over time on individual devices and at different rates.

\subsection{FL Retraining Framework}
\begin{figure*}[t]
  \centering
  \includegraphics[width=\linewidth]{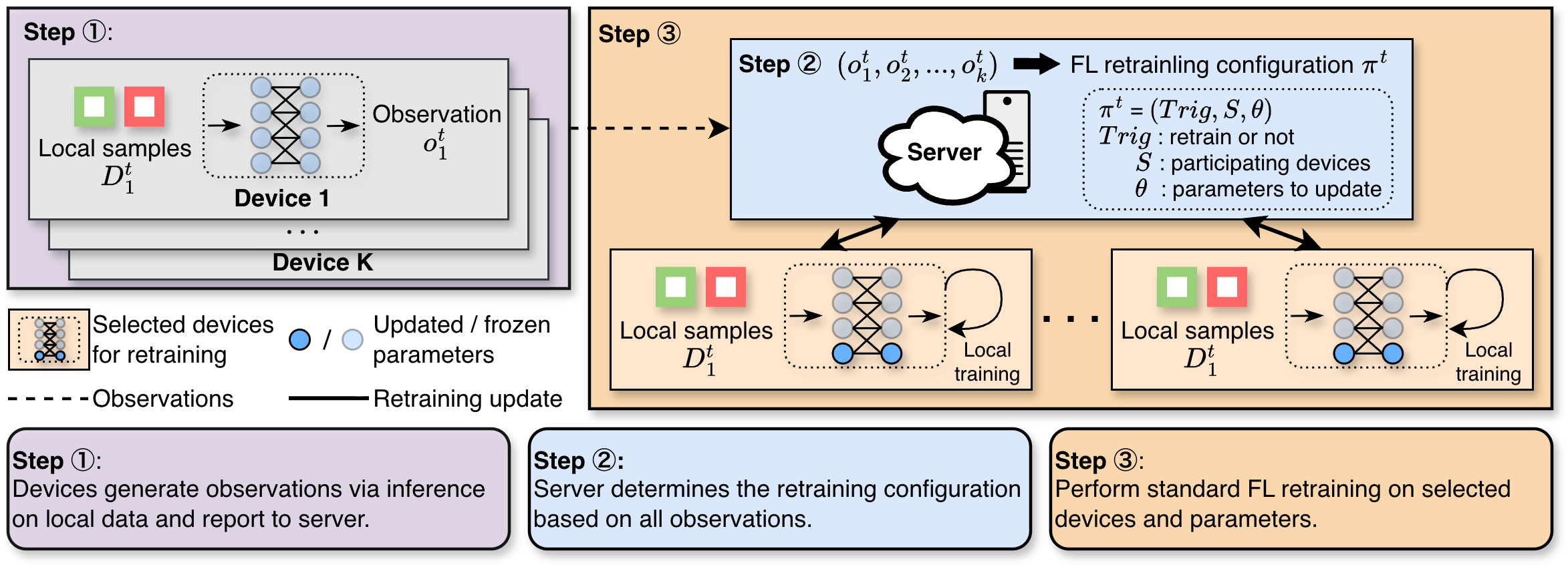}
  \caption{
    The three steps in the \ac{FL} retraining framework under asynchronous data drift. \textit{Step 1:} At each time step $t$, devices perform local inference and report observations to the server. \textit{Step 2:} The server determines the retraining configuration $\pi^t=(Trig,S,\theta)$, which specifies whether to retrain ($Trig$), which devices participate ($S$), and which parameters to update ($\theta$). \textit{Step 3:} If retraining is triggered, the selected devices perform FL retraining on the specified parameters.
  }
  \label{fig:fcl_framework}
\end{figure*}

To mitigate asynchronous data drift in FL, the server needs to monitor local inference statistics (e.g., accuracy) and decide whether to start FL retraining. Figure~\ref{fig:fcl_framework} presents the three steps of the retraining framework in an FL system. 

\textit{Step 1}: At each time step $t$, the server collects observation ${o}_c^{t}$ on local inference from all devices. The inference observation ${o}_c^{t}$ includes measured values, such as local model accuracy on newly generated data $\mathcal{D}_c^{t}$ at time step $t$.

\textit{Step 2}: Based on the inference observations 
$\{o_c^{t}\}_{c\in\mathcal{C}}$ from all devices, the server determines a retraining configuration $\pi^{t}$ that specifies whether retraining is performed, which devices participate, and which model parameters are updated. 
Formally, it can be expressed as
\begin{equation}
\pi^{t} = (\mathrm{Trig},\, \mathcal{S},\, \theta)
\end{equation}
where $\mathrm{Trig} \in \{0,1\}$ indicates whether retraining is triggered at time step $t$. $\mathcal{S} \subseteq \mathcal{C}$ denotes the set of devices selected to participate in FL retraining, and 
$\theta \in \Theta$ represents the subset of global model parameters $\Theta$ to be updated during retraining.

\textit{Step 3}: 
If FL retraining is triggered at time step $t$, the FL system performs standard FL retraining for $R$ rounds on the selected subset of devices $\mathcal{S}$. 
Each selected device $c \in \mathcal{S}$ performs local training using its newly generated local dataset $\mathcal{D}_c^{t}$ and updates the selected model parameters $\theta$. 
After completing the $R$ rounds of FL retraining, the server distributes the updated parameters to all devices.

\subsection{FL Retraining Costs}
\label{sec:prob-def}
Given the retraining configuration $(\mathrm{Trig}^t, \mathcal S^t, \theta^t)$ at time step $t$, the computational cost of performing $R$ rounds of FL retraining ($\mathrm{Trig}^t = 1$) is defined as
\begin{equation}
    \mathrm{Cost}(\mathcal S^t,\theta^t)
    \coloneqq
    \sum_{r=1}^{R}\sum_{c\in\mathcal S^t}
    \kappa_c(\theta^t,\mathcal D_c^t),
\end{equation}
where $\kappa_c$ denotes the computation overhead of local training on device $c$ using its most recent data $\mathcal D_c^t$ and model parameters $\theta^t$.

The total retraining cost over $T$ time steps is therefore
\begin{equation}
    \label{eq:totalcost}
    \mathrm{TotalCost}
    \coloneqq
    \sum_{t=1}^{T}\mathrm{Trig}^t \cdot 
    \mathrm{Cost}(\mathcal S^t,\theta^t).
\end{equation}

Equation~\ref{eq:totalcost} characterizes the relationship between the total retraining cost over time and the retraining configurations $\pi^t$ across $T$ time steps.
Triggering retraining at more time steps or having more devices participate in each retraining leads to a higher computational overhead.
In addition, the model parameters $\theta^t$ determine the number of parameters that needs to be updated, as more parameters will incur higher local training costs. 
As a result, the set of retraining configurations $\{\pi^t\}_{t=1}^{T}$ determines the total retraining cost over $T$ time steps under asynchronous data drift.

\subsection{Optimization Goal}
The retraining configuration not only affects the total retraining cost but also model accuracy. At time step $t$, we consider the average accuracy over $K$ devices after retraining under configuration $\pi^t$ as
\begin{equation}
    A_{\pi}^t \coloneqq \frac{1}{K}\sum_{c\in\mathcal C}\mathrm{Accuracy}_{c,\pi}^t,
\end{equation}
where $\mathrm{Accuracy}_{c,\pi}^t$ denotes the on-device accuracy measured after retraining under configuration $\pi^t$. The average accuracy over $T$ time steps is defined as
\begin{equation}
    \bar{A}(\pi) \coloneqq \frac{1}{T}\sum_{t=1}^{T} A_{\pi}^t.
\end{equation}

The goal of an FL system under asynchronous data drift is to optimize a sequence of retraining configurations, denoted as $\boldsymbol{\Pi} = \{\pi^t\}_{t=1}^{T}$, so as to maximize the accuracy achieved per unit retraining cost:
\begin{equation}
    \label{eq:goal}
    \boldsymbol{\Pi}^*
    = \operatorname*{arg\,max}_{\boldsymbol{\Pi}}
    \frac{\bar{A}(\boldsymbol{\Pi})}{\mathrm{TotalCost}(\boldsymbol{\Pi})}.
\end{equation}
        
Equation~\ref{eq:goal} provides a unified formulation of existing retraining strategies under asynchronous data drift. For example, classic FCL methods~\cite{kirkpatrick2017ewc} adopt a retraining configuration in which retraining is triggered based on global model accuracy, all devices participate in each retraining round (i.e., $\mathcal S^t = \mathcal C$), and the parameters of the entire model are updated. While such a strategy achieves good accuracy, it incurs a high computational cost. Clustering-based methods trigger retraining when a group’s average accuracy falls below a threshold and update only group-specific parameters.

\subsection{Challenges in Optimizing Accuracy-Cost Trade-off under Asynchronous Data Drift}
Under asynchronous data drift, different devices may exhibit different drift rates. Consequently, at time step $t$, only a subset of devices may experience accuracy degradation. If the server triggers retraining whenever any device falls below a predefined accuracy threshold, then retraining may inevitably occur frequently, incurring substantial computational costs of FL retraining on all devices. In contrast, if retraining is delayed until a large fraction of devices drop below the accuracy threshold, then devices that have already experienced drift may suffer from poor performance persistently. The tension between timely retraining and retraining cost makes optimizing Equation~\ref{eq:goal} challenging under asynchronous data drift.

%% file: sections/method.tex
In this section, we present the design of \method. 
We first describe its design principles, followed by the MoE-based architecture used in \method. 
Based on this architecture, we introduce two key techniques in \method.
First, we describe how the server clusters devices with similar data distributions using local gating outputs. Then, we present the generation of global and group retraining configurations, specifying whether to retrain, which devices participate, and which parameters to update.
Finally, we summarize the end-to-end pipeline of \method.

\subsection{Design Principle of \method}
\label{sec:sys:design}
To optimize Equation~\ref{eq:goal} under asynchronous data drift, \method aims to maintain a high average accuracy across all devices over time while reducing retraining costs. To this end, \method adopts an MoE-based architecture and leverages it to decompose FL retraining into:
(i) \textit{global retraining} of shared parameters that capture globally transferable knowledge across all devices, and
(ii) \textit{group retraining} of local parameters that adapt to the data distribution of each device group. Here, a group refers to a subset of devices that share similar data distributions at a given time step.

By decoupling traditional FL retraining in this way, \method can retrain only a small group of devices whenever they experience accuracy degradation, without incurring the cost of retraining on all devices. Meanwhile, retraining with all devices is triggered only when the server detects that the shared parameters do not adequately capture the globally transferable knowledge. In addition, in both cases, whether retraining local or global parameters, only a subset of parameters will be updated, further reducing the overall retraining cost.

\begin{figure}[t]
  \centering
  \includegraphics[width=\linewidth]{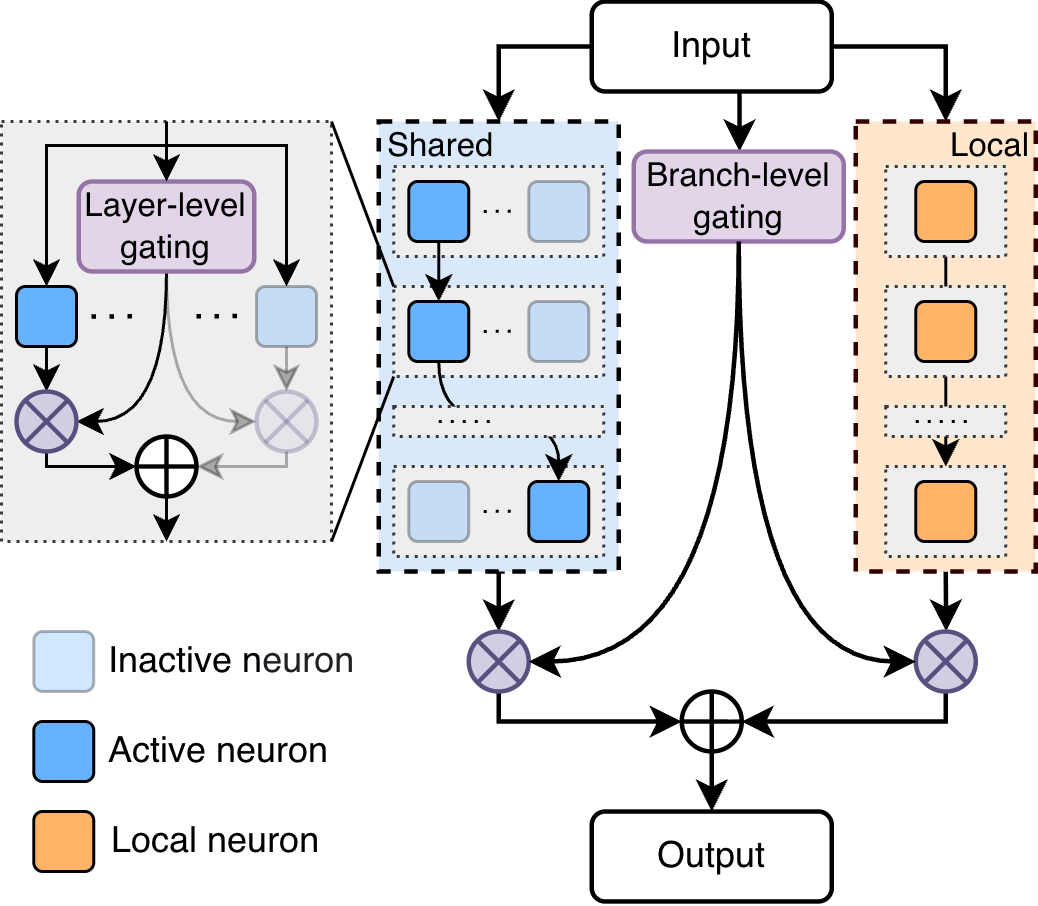}
  \caption{
      The MoE-based architecture in \method. 
      A branch-level soft gating network decomposes the model into a shared branch (blue) and a local branch (orange), containing shared and local parameters respectively. 
      Within the shared branch, a layer-level hard gating network activates different subsets of neurons for inputs from different distributions.
   }
  \label{fig:moe}
\end{figure}

\subsection{Mixture-of-Experts Architecture in \method}
\label{sec:sys:moe}
An MoE architecture is a modular neural network in which multiple specialized sub-networks, referred to as \emph{experts}, are coordinated by a gating network~\cite{shazeer2017moe}.
For each input, the gating network dynamically selects a small subset of experts and combines their outputs to produce the final output. This design allows the model to scale capacity during training while incurring only a small increase in inference cost~\cite{dai2024deepseekmoe}.

Figure~\ref{fig:moe} illustrates how the MoE architecture is used in \method. Specifically, MoE architecture is adopted at two levels.
First, at the branch-level, \method decomposes the on-device model into two branches: (i) one containing shared parameters (shown in blue), updated during global retraining when the globally transferable knowledge becomes outdated, and (ii) the other containing local parameters (shown in orange), updated during group retraining to adapt to data distribution shifts within specific device groups. Shared parameters are updated to handle global data drift across all devices, while local parameters are trained to adapt to asynchronous data drift of each group.
Second, in the shared-parameter branch, each layer comprises a gating module that controls how the outputs of different experts in that layer are combined.
At the branch-level, \method uses soft gating, where the gate outputs continuous scores to weight different branches. In contrast, at the layer-level, \method uses hard gating with binary decisions (0 or 1) to select active neurons. 

The branch-level gating network enables a separation between global and local parameters. As a result, \method can adapt to data shift in a group by selectively updating local parameters, and to global data shift by updating shared parameters. 
In addition, the layer-level gating network enables the server to cluster devices with similar local data distributions into groups, without requiring devices to send user input data to the server (see Section~\ref{sec:sys:clustering}).
This architecture enables \method to selectively update only the relevant subset of parameters when adapting to asynchronous data drift, rather than updating all parameters, thereby reducing the overall retraining cost.

\subsection{Device Clustering}
\label{sec:sys:clustering}

\subsubsection{Inference on Devices Prior to Retraining} At each time step $t$, every device checks for local data drift on its recent data. Specifically, each device performs local inference on its recent data samples and records the inference outputs to send to the server. We refer to these inference records as the observation $o_c$ of device $c$. Each record includes the device’s average local accuracy and an aggregated gating matrix that summarizes the outputs (activations) patterns of the layer-level gating network.  
The aggregated gating matrix is computed by averaging the gate activations across samples for each class, weighted by soft labels. 
This matrix captures how neurons in the shared branch respond to the device's local data, reflecting the underlying data distribution of recent samples.

Table~\ref{fig:obs} provides an example observation on device $c$. The observations from all devices are sent to the server, which cluster devices into groups based on the similarity of their aggregated gating matrices and then generates retraining configurations. 

\begin{table}[t]
\centering
\caption{
    Example on-Device observation $o_c^t$
}
\label{fig:obs}
    \renewcommand{\arraystretch}{1.5}
    \begin{tabularx}{\linewidth}{llX}
        \toprule
        \textbf{Component Observed} & \textbf{Sample Value} & \textbf{Description} \\
        \midrule
        Average Local Accuracy &
        $0.60$  &
        Accuracy on recent local data
        \\
        
        Aggregated Gating Matrix & 
            $\begin{bmatrix} 
            0.1 & \cdots & 0.9 \\ 
            0.4   &  \cdots & 0.6 \\
            \vdots & \ddots & \vdots
            \end{bmatrix}$ & 
            \vspace{-2.75em}
            Average of gate activations on recent local data
            \\
        \bottomrule
    \end{tabularx}
\end{table}

\subsubsection{Clustering Devices on the Server}
After receiving observations from all devices, the server clusters devices into groups based on the similarity of their aggregated gating matrices, so that devices with similar underlying data distributions are assigned to the same group.
Specifically, in \method, the agglomerative hierarchical clustering~\cite{murtagh2012ahc} algorithm is employed.
It determines the number of groups automatically via a distance threshold.  
In addition, we enforce a minimum group size: if any group contains fewer than a predefined minimum number of devices, it is merged with the nearest group based on the similarity of their aggregated gating matrices.
The clustering results will be used to determine which devices should participate in group retraining. Devices within the same group share group-specific local parameters.

\subsection{Generating Retraining Configuration}
\label{sec:sys:retraining}
After clustering, the server generates retraining configurations to determine: \textit{(i) whether retraining is needed},
\textit{(ii) if so, whether it should be triggered across all devices or within specific groups}, and 
\textit{(iii) which parameters should be updated}. 
In particular, \method generates two types of retraining configurations: \textit{Global retraining} and \textit{Group retraining}.

\subsubsection{Global Retraining of Shared Parameters}
The server first considers generating a global retraining configuration
$\pi_{g} = (\mathrm{Trig},\, \mathcal{S},\, \theta)$.

\textit{(i) Trigger decision $\mathrm{Trig}$.}
The decision for when global retraining is triggered is based on the average accuracy across all devices. 
When the average accuracy falls below a predefined threshold, it indicates a significant shift in the overall data distribution across devices, suggesting that the globally transferable knowledge captured by the shared parameters has become outdated. 
Accordingly, global retraining is triggered to update the shared parameters.

\textit{(ii) Participating devices $\mathcal{S}$.}
All devices participate in global retraining, i.e.,
$\mathcal{S} = \{1,2,\dots,K\}$.

\textit{(iii) Retrained parameters $\theta$.}
Only the shared parameters and branch-level gating parameters are updated, denoted as $\theta_{\text{shared}}$. The local parameters remain frozen.

\subsubsection{Group Retraining of Local Parameters}  
The server also generates group retraining configurations for each group,
$\pi_{g_i} = (\mathrm{Trig},\, \mathcal{S},\, \theta)$.

\textit{(i) Trigger decision $\mathrm{Trig}$.}  
The decision for group retraining is based on whether accuracy degradation is observed within a group. Specifically, if the average local inference accuracy of devices in a group falls below a predefined threshold, partial retraining is triggered for that group. This strategy reduces accuracy degradation within local groups.

\textit{(ii) Participating devices $\mathcal{S}$.}  
Let $\mathcal{G}_j$ denote the set of devices in group $j$.  
If group $j$ is detected by the server to have accuracy degradation, then only devices in this group participate in retraining, i.e.,
\[
\mathcal{S} = \mathcal{G}_j .
\]

\textit{(iii) Retrained parameters $\theta$.}   
During group partial retraining for group $j$, only the local parameters of this group $\theta_{\text{local}}^{j}$ and branch-level gating parameters are updated, while the shared parameters $\theta_\text{shared}$ remain frozen.

\subsection{\method Pipeline}
\label{sec:sys:pipeline}
\begin{figure*}[htb]
  \centering
  \includegraphics[width=\linewidth]{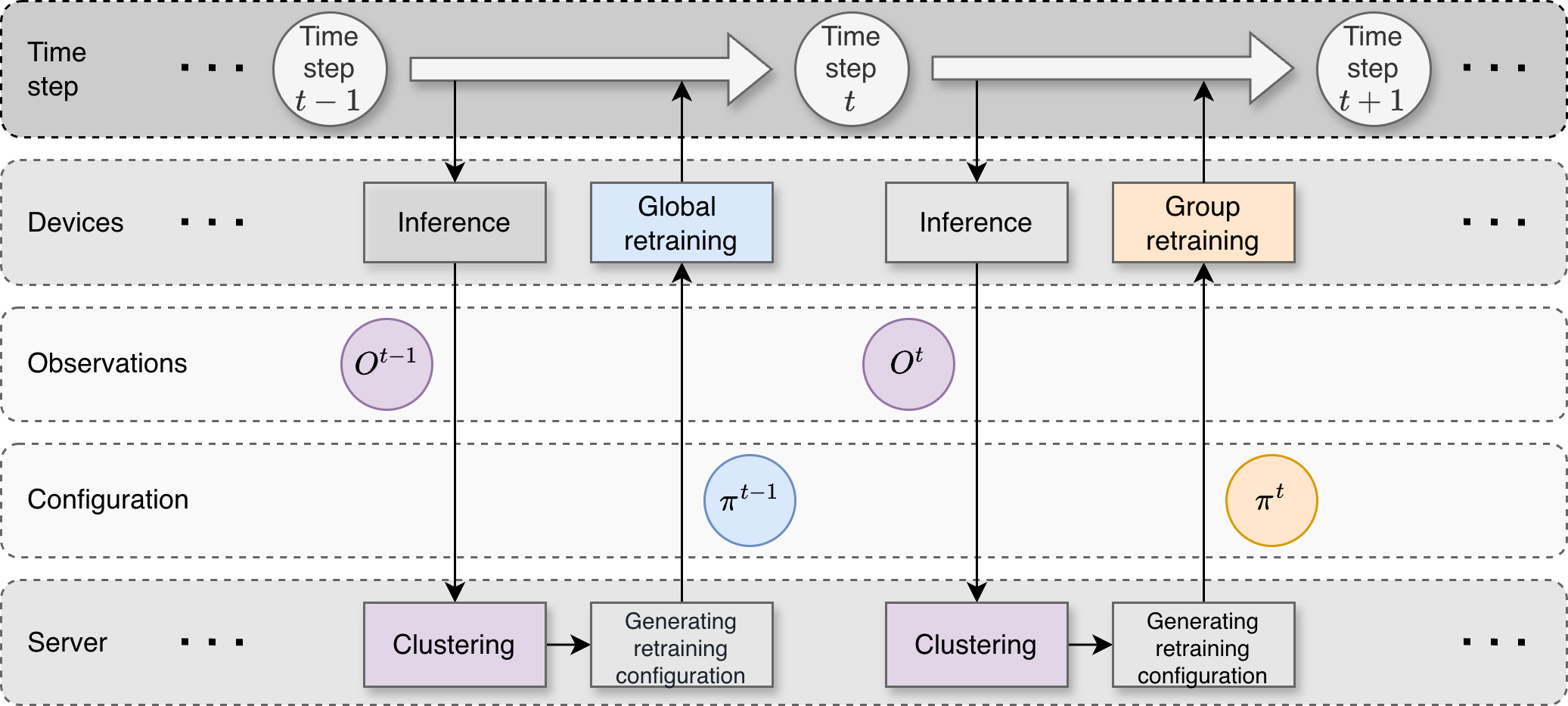}
  \caption{
    The integrated pipeline of \method.
    At each time step $t$, the server collects the set of observations $O^t=\{o_c^t\}$ via inference from all devices. 
    Based on $O^t$, the server performs clustering and generates the retraining configuration $\pi^t$.
    The pipeline dynamically triggers either \textit{global retraining} of shared parameters or \textit{group retraining} of local parameters in specific groups.
  }
  \label{fig:timeline}
\end{figure*}

Figure~\ref{fig:timeline} shows the integrated pipeline of \method for adapting to asynchronous data drift.
At each time step ($t$), \method follows a four-step pipeline: inference on the devices prior to retraining, clustering of devices, generating retraining configuration, and FL retraining.

\textit{Step 1: Inference} -- Each device performs local inference on a batch of recent samples and reports its observation to the server.

\textit{Step 2: Clustering} -- Based on the observations from all devices, the server clusters devices with similar gating matrix from the observations into groups. This clustering is performed before any retraining configuration is generated.

\textit{Step 3: Retraining configuration} -- Using the observations and clustering results, the server determines whether retraining is needed. If a global shift in the shared parameters is detected, global retraining is triggered. If accuracy degradation is detected in specific groups, group retraining is triggered. Global retraining and group retraining are not mutually exclusive and can be triggered simultaneously within the same time step.

\textit{Step 4: FL retraining} --
Based on the retraining configurations, the participating devices execute the specified retraining strategies.
After retraining, the updated parameters are aggregated and redistributed to devices.

In summary, \method combines an MoE-based architecture with decouple global and group retraining to handle asynchronous data drift. By clustering devices and selectively updating shared or group-specific parameters, it maintains high accuracy while reducing retraining cost.

%% file: sections/evaluation.tex
In this section, we evaluate \method under an asynchronous data drift setting and compare it against five state-of-the-art baselines spanning standard FCL method, personalized FL, and clustering-based approaches.  
We first describe the experimental setup, including the model-dataset configurations, the simulation of asynchronous data drift, and the baseline methods. 
We then present the evaluation results, including both theoretical cost analysis and real-world latency measurements on a Raspberry Pi-based testbed, along with an impact on key hyperparameters.
Our two takeaways from the evaluation are:

1) \textbf{Higher accuracy per retraining cost:} across all six experimental configurations we considered, \method achieves the highest or comparable accuracy to the strongest baselines while reducing the total retraining cost (floating-point operations, FLOPs) by up to \costredT. This leads to the highest accuracy per retraining cost ($\mathcal{E}$), up to 2.3$\times$ higher than the strongest baseline.

2) \textbf{End-to-end performance on a real-world IoT prototype:} On a real-world IoT prototype, \method achieves the highest accuracy across all experimental settings while reducing retraining time by up to 20\%, resulting in the highest $\mathcal{E}$, up to 1.2$\times$ higher than the strongest baseline.

\subsection{Experimental Setup}
\label{sec:eva:setup}

\textbf{Model and dataset:}
we evaluate \method using three model–dataset pairs: cResNet-S and cViT-S on DG5, cResNet-M and cViT-M on PACS and DomainNet. Specifically, we consider two base architectures: ResNet~\cite{he2016resnet}, and ViT~\cite{dosovitskiy2021vit}, covering two widely used model architectures---convolutional neural networks and vision transformers.
Each base architecture is adapted into two scale variants, small (S) and medium (M), by configuring different layer depths and model dimensions.
To support the MoE-based architecture of \method, we augment each base model with an MoE, resulting in four MoE variants: cResNet-S, cResNet-M, cViT-S, and cViT-M. Table~\ref{tab:model_size} summarizes the configurations of these variants.

\begin{table}[t]
\centering
\caption{
    Models used in the evaluation of \method. For ResNet-based variants, \textit{layers} lists the number of blocks per stage, \textit{experts} is the number of expert sub-networks per layer in the shared branch, and \textit{top-k} is the number of experts activated per input by the layer-level gating. For ViT-based variants, \textit{patch} is the input patch size, \textit{dim} is the feature dimension, \textit{depth} is the number of transformer layers, and \textit{heads} is the number of attention heads.
}
\label{tab:model_size}
\begin{tabularx}{\linewidth}{l X X l }
    \toprule
    \textbf{Model} & \textbf{Base} & \textbf{Size} & \textbf{Architecture configuration} \\
    \midrule
    cResNet-S & ResNet & 4.8M & layers=[1,1,1], experts=3, top-k=1 \\
    cResNet-M & ResNet & 20.6M & layers=[2,2,1,1], experts=3, top-k=1 \\
    cViT-S & ViT & 2.0M & patch=4, dim=128, depth=6, heads=4 \\
    cViT-M & ViT & 9.1M & patch=16, dim=192, depth=12, heads=3 \\
    \bottomrule
\end{tabularx}
\end{table}

The above models are evaluated on three multi-domain image classification datasets that are widely used in the continual learning literature.
(i) DG5~\cite{cho2023dg5, sun2023dg5} consists of handwritten digit images collected from five domains: MNIST, MNIST-M, SVHN, SYN, and USPS.
(ii) PACS~\cite{li2017pacs, sun2025pacs-domainnet} is a multi-domain object recognition dataset comprising images from four domains: Photo, Art Painting, Cartoon, and Sketch.
(iii) DomainNet~\cite{peng2019domainnet, sun2025pacs-domainnet} is a curated subset of the original DomainNet dataset, designed to model domain drift across visual styles. It consists of six domains: Real, Clipart, Painting, Sketch, Infograph, and Quickdraw, each exhibiting distinct distributional characteristics in terms of texture, abstraction level, and drawing style.
Table~\ref{tab:datasets} summarizes the number of domains, classes, and samples in the datasets.

\textbf{Metrics:}
we compare different retraining policies using the optimization objective formulated in Equation~\ref{eq:goal}, 
which measures the accuracy achieved per retraining cost, $\boldsymbol{\mathcal{E}}$. 
In addition, we report the individual components of $\boldsymbol{\mathcal{E}}$, including $\boldsymbol{\bar{A}}$, 
the average accuracy across all devices over all time steps, and \textbf{TC}, the cumulative retraining cost measured by the total computational overhead in FLOPs.
An ideal retraining policy maximizes $\boldsymbol{\mathcal{E}}$ while jointly improving $\boldsymbol{\bar{A}}$ and reducing \textbf{TC}.

\textbf{Hardware:} 
experiments are conducted on a machine with two NVIDIA RTX A6000 GPUs (48GB memory each) and an AMD EPYC 7713P 64-Core Processor.
For evaluation on a real-world IoT prototype, we deploy \method on 20 Raspberry Pi 4 devices.

\begin{table}[t]
\centering
\caption{Characteristics of the datasets.}
\label{tab:datasets}
    \begin{tabularx}{\linewidth}{XYYY}
    \toprule
    \textbf{Dataset} & \textbf{\#Domains} & \textbf{\#Classes} & \textbf{\#Images} \\
    \midrule
    DG5 & 5 & 10 & 215,695 \\
    PACS & 4 & 7 & 9,991 \\
    DomainNet & 6 & 7 & 16,824 \\
    \bottomrule
    \end{tabularx}
\end{table}

\textbf{FL retraining hyperparameters:}
at each FL retraining time step, we perform five rounds of federated aggregation using the standard FedAvg algorithm~\cite{mcmahan2017FedAvg}. 
For each device, local training is conducted for up to 20 epochs per round with a batch size of 8 and a learning rate of 0.001. 
We additionally adopt an early stopping strategy~\cite{prechelt1998early_stop} with a patience of 5 epochs to prevent overfitting during local training.

\textbf{Creating asynchronous data drift:}
we simulate asynchronous data drift through a four-step process.

\textit{(i) Time horizon and data arrival} :
We simulate an FL system consisting of a central server and 20 devices over 30 time steps under asynchronous data drift.
Each time step represents a potential data drift event for each device.
At each time step, each device receives a local dataset of 30 images sampled from the dataset based on its current underlying data distribution, representing the newly arrived local data distribution at that time step. Of these, 20 images are used for retraining together with the 10 validation images retained from the previous time step, while the remaining 10 are used as a local validation set for monitoring accuracy. At the first time step, only the 20 images are used for initial training.

\textit{(ii) Creating drift events}: 
At each time step, we independently determine whether a data drift event occurs on each device.
Specifically, 10\%--15\% of the devices (i.e., 2 to 3 devices) are randomly selected to undergo a new data drift event,
while the remaining devices retain their current data distributions.

\textit{(iii) Assigning drift patterns}: 
For each selected device, we assign one of the following drift patterns:
(1) \emph{Instantaneous drift}, where the device's underlying data distribution immediately and completely shifts to a new randomly selected domain. All subsequent local data is then sampled from this new domain.
(2) \emph{Incremental drift}, where the device’s data distribution gradually shifts toward a new domain over a period of up to 4 consecutive time steps (15\% of the total time steps).
The cap of 15\% is chosen to ensure enough time to simulate a gradual shift. 
During the incremental drift period, local data is sampled from a mixture of the original and target domains,
with the proportion of data drawn from the target domain increasing linearly at each time step, until the device's distribution has fully shifted to the target domain.

\textit{(iv) Updating and sampling the local distribution}: 
At each time step, each device first updates its local data distribution by applying all drift events, including both ongoing incremental drift processes and new drift events.
Data samples are then drawn from this updated distribution.
Note that when multiple drift events overlap on the same device, they are applied sequentially in the order in which they were triggered.

\textbf{Baselines:}
we compare \method against five baselines discussed below.
Each baseline is characterized by its retraining configuration
$\Pi = (Trig, S, \theta)$, as defined in Section~\ref{sec:model},
where $Trig$ denotes the retraining trigger, $S$ the set of participating devices,
and $\theta$ the set of parameters to be retrained.

(i) \textit{FCL-AveTrig} -- a standard FCL retraining pipeline~\cite{yoon2021FedWeIT} is considered in which retraining is triggered when the average accuracy across all devices falls below a threshold,
after which retraining is executed on all devices and all model parameters are updated.

(ii) \textit{FCL-perDevice} -- similar to the above FCL pipeline but retraining is triggered when any device’s accuracy falls below a threshold,
and retraining is executed only on the devices where accuracy falls below the threshold, while still updating all model parameters.

We further evaluate two personalized federated learning (PFL) baselines~\cite{yi2024moe}, which have personalized parameters that adapt to each device's local data distribution.

(iii) \textit{PFL-AveTrig} -- retraining is triggered when the average accuracy across all devices falls below a threshold and is executed on all devices, while both shared and local parameters are updated, with global averaging applied only to the shared parameters.

(iv) \textit{PFL-perDevice} -- retraining is triggered when any device’s accuracy falls below a threshold and is executed only on the devices where accuracy falls below the threshold, where both shared and local parameters are updated, with global averaging applied only to the shared parameters.

Finally, we include a cluster-based FL baseline~\cite{jothimurugesan2023feddrift,peng2025feddaa},
which clusters devices into groups and maintains a separate model for each group.

(v) \textit{Cluster-based} -- retraining is triggered when the average accuracy of a group falls below a threshold,
retraining is executed on devices within the group,
and all parameters are updated.

The above baselines cover standard FCL and state-of-the-art MoE-based PFL methods under two different trigger conditions. In addition, a cluster-based FL approach is also included.
All baselines use the same underlying model architecture for FL retraining to ensure a fair comparison.


\subsection{Comparison of \method with Baselines}
\label{sec:eva:result}

We evaluate \method against baselines in terms of accuracy per retraining cost, accuracy preservation under data drift, and accumulated retraining cost.

\paragraph{Higher accuracy per retraining cost}
We define an efficiency metric $\mathcal{E}$ as the mean accuracy across all steps divided by the total retraining cost. $\mathcal{E}$ reflects the accuracy achieved per retraining cost under data drift; a higher $\mathcal{E}$ indicates a more efficient method for mitigating data drift. Table~\ref{tab:main_results} presents $\mathcal{E}$ of \method and other baselines across all six experimental configurations. In general, \method achieves the highest $\mathcal{E}$ in five of the six configurations.

On the DG5 dataset, when training cResNet-S, \method achieves \(1.3\times\)--\(3.8\times\) higher \(\mathcal{E}\) than other baselines. 
For cViT-S, however, \method does not achieve the highest \(\mathcal{E}\). 
We conjecture that, under this configuration, \method exhibits retraining behavior similar to FCL-AveTrig and PFL-AveTrig, where global retraining is frequently triggered due to the limited capacity of cViT-S~\cite{dosovitskiy2021vit}.
It is worth noting that, although PFL-perDevice and FCL-perDevice achieve higher \(\mathcal{E}\) than \method when training cViT-S on DG5, this efficiency gain comes at the expense of accuracy, which decreases by 1\% and 2\%, respectively.

On PACS and DomainNet, which exhibit stronger domain shifts than DG5, \method consistently achieves the highest \(\mathcal{E}\) among all baselines. 
On PACS, \method outperforms the strongest baseline by \(2.3\times\) for cResNet-M, where the best baseline is PFL-perDevice. For cViT-M, \method surpasses the strongest baseline, PFL-AveTrig, by \(1.3\times\).
On DomainNet, \method outperforms the strongest baseline, PFL-AveTrig, by \(1.6\times\) for cResNet-M and by \(2.3\times\) for cViT-M.
Moreover, on both PACS and DomainNet, across cResNet-M and cViT-M models, \method achieves the highest accuracy while incurring \costredc lower cumulative retraining cost than other baselines.

\begin{table*}[h]
    \caption{Efficiency comparison across datasets and models. Each cell reports $\mathcal{E}$ ($\bar{A}$/TC), where $\mathcal{E} = \bar{A}/\text{TC}$ measures accuracy achieved per unit retraining cost. * marks the best result for each experimental configuration.}
    \label{tab:main_results}
    \centering
    \begin{tabularx}{\textwidth}{
      X          
      C{2.35cm}C{2.35cm}  
      C{2.175cm}C{2.18cm}  
      C{2.175cm}C{2.18cm}  
    }
        \toprule
        \multirow{2}{*}{\textbf{Method}} 
        & \multicolumn{2}{c}{\textbf{DG5}}
        & \multicolumn{2}{c}{\textbf{PACS}}
        & \multicolumn{2}{c}{\textbf{DomainNet}} \\
        \cmidrule(lr){2-3}
        \cmidrule(lr){4-5}
        \cmidrule(lr){6-7}
        & \textbf{cResNet-S} & \textbf{cViT-S}
        & \textbf{cResNet-M} & \textbf{cViT-M}
        & \textbf{cResNet-M} & \textbf{cViT-M} \\
        \midrule
        \rowcolor{cyan!20}
        DriftGuard 
        	& \textit{12.14} \text{(0.85 / 0.07)}*
        	& \textit{7.44} \text{(0.67 / 0.09)} 
        	& \textit{3.75} \text{(0.60 / 0.16)}*
        	& \textit{3.47} \text{(0.59 / 0.17)}*
        	& \textit{3.81} \text{(0.61 / 0.16)}*
        	& \textit{8.57} \text{(0.60 / 0.07)}*\\
        FCL-AveTrig 
        	& \textit{3.19} \text{(0.86 / 0.27)} 
        	& \textit{5.67} \text{(0.68 / 0.12)} 
        	& \textit{1.11} \text{(0.60 / 0.54)} 
        	& \textit{2.03} \text{(0.59 / 0.29)} 
        	& \textit{2.10} \text{(0.61 / 0.29)} 
        	& \textit{2.86} \text{(0.60 / 0.21)} \\
        FCL-perDevice 
        	& \textit{7.82} \text{(0.86 / 0.11)} 
        	& \textit{9.29} \text{(0.65 / 0.07)}  
        	& \textit{1.25} \text{(0.60 / 0.48)} 
        	& \textit{2.19} \text{(0.57 / 0.26)} 
        	& \textit{1.42} \text{(0.61 / 0.43)} 
        	& \textit{2.32} \text{(0.58 / 0.25)} \\
        PFL-AveTrig 
        	& \textit{4.05} \text{(0.85 / 0.21)} 
        	& \textit{6.80} \text{(0.68 / 0.10)} 
        	& \textit{1.28} \text{(0.60 / 0.47)} 
        	& \textit{2.68} \text{(0.59 / 0.22)} 
        	& \textit{2.44} \text{(0.61 / 0.25)} 
        	& \textit{3.75} \text{(0.60 / 0.16)} \\
        PFL-perDevice 
        	& \textit{9.56} \text{(0.86 / 0.09)}
        	& \textit{11.00} \text{(0.66 / 0.06)}*
        	& \textit{1.62} \text{(0.60 / 0.37)} 
        	& \textit{2.50} \text{(0.55 / 0.22)} 
        	& \textit{1.76} \text{(0.60 / 0.34)} 
        	& \textit{2.90} \text{(0.58 / 0.20)} \\
        Cluster-based 
        	& \textit{4.10} \text{(0.86 / 0.21)} 
        	& \textit{7.25} \text{(0.58 / 0.08)} 
        	& \textit{1.40} \text{(0.60 / 0.43)} 
        	& \textit{1.22} \text{(0.49 / 0.40)} 
        	& \textit{2.03} \text{(0.61 / 0.30)} 
        	& \textit{1.27} \text{(0.52 / 0.41)} \\
    \bottomrule
    \end{tabularx}
\end{table*}

\begin{figure*}[htb]
  \centering
  \includegraphics[width=\linewidth]{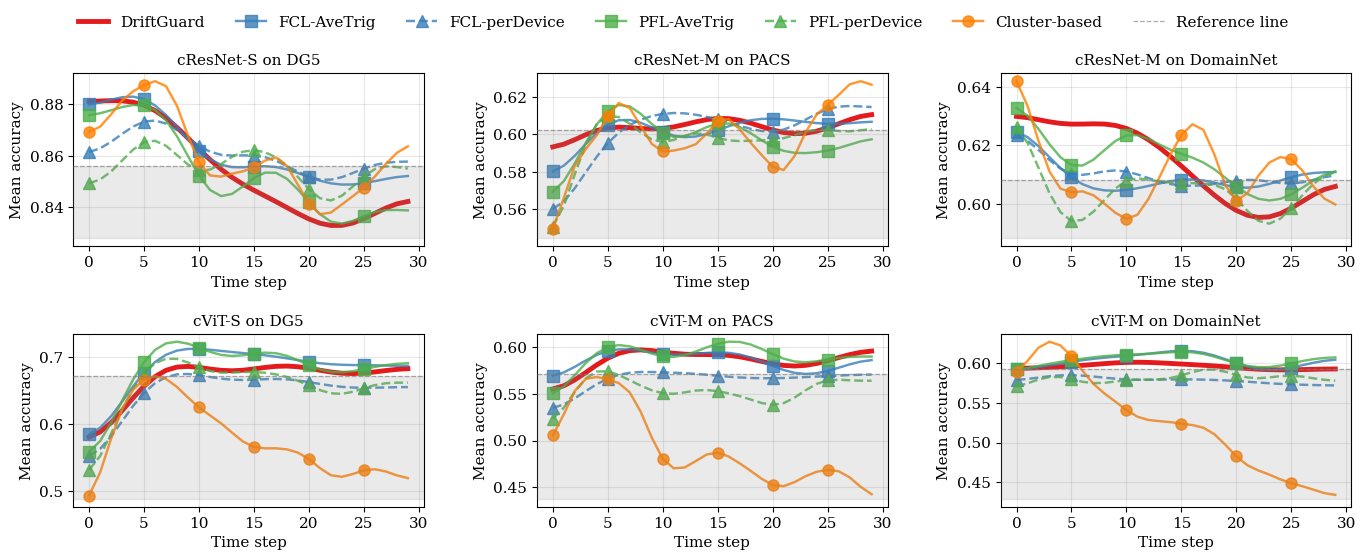}
  \caption{
      Mean accuracy ($\bar{A}$) over time steps across datasets and models, with curves smoothed using a Gaussian filter ($\sigma = 1.5$). The dashed grey line denotes the reference threshold, defined as the median accuracy across all methods and time steps. The shaded region indicates the portion below the reference threshold.
  }
  
  \label{fig:mean_acc}
\end{figure*}

\paragraph{Accuracy preservation under data drift}
Figure~\ref{fig:mean_acc} further presents the accuracy trajectories of \method and other baselines across the six experimental configurations. To better compare how well each method preserves accuracy over time steps, we include a reference line for comparison. This line denotes the median accuracy of all methods over all 30 time steps. In addition, Table~\ref{tab:maintenance} reports the total number of time steps for which each method remains above the reference accuracy.
\method achieves the highest or a comparable count to the strongest baseline across all configurations on PACS and DomainNet, demonstrating its ability to preserve accuracy under asynchronous drift.
For example, on PACS, \method maintains accuracy above the reference threshold for 16 time steps with cResNet-M, only 2 fewer than the strongest baseline, FCL-perDevice. With cViT-M, \method stays above the reference threshold for 18 time steps, just 1 fewer than the strongest baseline, PFL-AveTrig. On DomainNet, \method remains above the reference threshold for 17 time steps with cResNet-M, the highest among all methods in this setting. With cViT-M, \method stays above the reference threshold for 21 time steps, 2 fewer than the strongest baseline.


\begin{table}[t]
    \caption{
    Number of time steps each method remains above the reference threshold (out of 30 total time steps).
    }
    \label{tab:maintenance}
    \setlength{\tabcolsep}{0.1pt}
    \centering
    \begin{tabularx}{\linewidth}{
      X          
      C{1.3cm}C{1cm}  
      C{1.35cm}C{1cm}  
      C{1.35cm}C{1cm}  
    }
        \toprule
        \multirow{2}{*}{\textbf{Method}} 
        & \multicolumn{2}{c}{\textbf{DG5}}
        & \multicolumn{2}{c}{\textbf{PACS}}
        & \multicolumn{2}{c}{\textbf{DomainNet}} \\
        \cmidrule(lr){2-3}
        \cmidrule(lr){4-5}
        \cmidrule(lr){6-7}
        & \textbf{cResNet-S} & \textbf{cViT-S}
        & \textbf{cResNet-M} & \textbf{cViT-M}
        & \textbf{cResNet-M} & \textbf{cViT-M} \\
        \midrule
        \rowcolor{cyan!20}
        DriftGuard 
        	& 13 & 19 & 16 & 21 & 17 & 21 \\
        FCL-AveTrig 
        	& 15 & 24 & 14 & 21 & 13 & 23 \\
        FCL-perDevice 
        	& 20 & 9 & 18 & 14 & 16 & 8 \\
        PFL-AveTrig 
            & 13 & 22 & 15 & 22 & 17 & 23 \\
        PFL-perDevice 
            & 14 & 14 & 17 & 8 & 13 & 9 \\
        Cluster-based 
            & 15 & 3 & 16 & 4 & 16 & 6 \\
        \bottomrule
    \end{tabularx}
\end{table}

\paragraph{Lower Accumulated Costs}

\begin{figure*}[h]
  \centering
  \includegraphics[width=\linewidth]{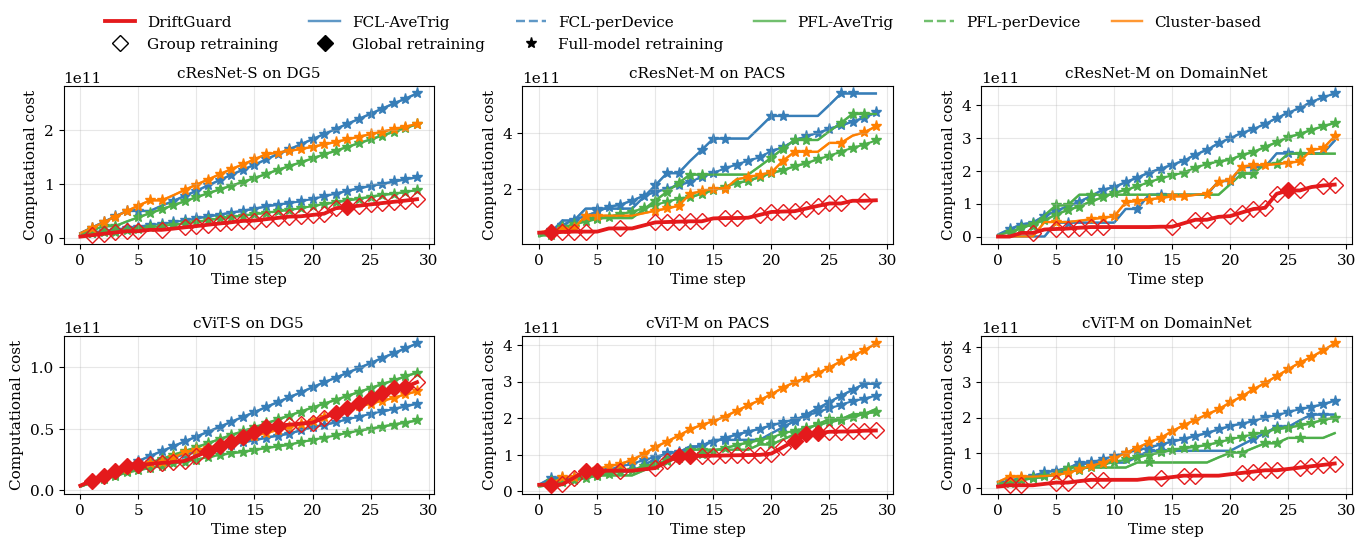}
    \caption{Cumulative retraining cost (TC) over time steps. Markers denote different types of retraining events.}
  \label{fig:cost}
\end{figure*}

Figure~\ref{fig:cost} shows the cumulative computational cost over time steps across all experimental configurations.
Markers denote retraining events, and different marker types indicate the corresponding retraining configurations.
Across all datasets, \method reduces the total retraining cost by \costred compared to the strongest baseline.

In detail, the retraining cost reduction of \method comes from two key factors: a lower cost per retraining event and a lower trigger frequency of global retraining.
First, group retraining majorly accounts for the majority of retraining events in \method (except when training cViT-S on DG5). Because each group retraining updates only the local parameters of a subset of devices, rather than full parameters across all devices, it incurs a much smaller increase in computation cost than the full-model retraining.
Second, \method triggers global retraining only when a significant drop in overall accuracy across devices is detected, which occurs only a few times over the 30 time steps. This is because most drift events are localized within groups and can be addressed by lightweight group retraining.
For example, on PACS, \method triggers global retraining only 1 time for cResNet-M, whereas the strongest baselines, FCL-AveTrig triggers global retraining 13 times. For cViT-M, \method triggers global retraining 8 times, whereas the strongest baselines, PFL-AveTrig, triggers global retraining 15 times.
Overall, \method reduces the number of global retraining triggers by 4 to 25 times steps over 30 time steps compared to other strongest baselines.

In summary, across all dataset-model configurations, \method achieves the highest or comparable accuracy to the strongest baselines while reducing the total retraining cost by up to \costredT.
As a result,\method attains the highest accuracy per retraining cost ($\mathcal{E}$) in five out of six configurations, up to \effimpt higher than the strongest baseline.

\subsection{Evaluation on a Real-World IoT Prototype}
\label{sec:eva:realdevice}

We further evaluate \method on a real-world IoT prototype. Specifically, we build a testbed with 20 Raspberry Pi 4 devices. We train cResNet-M and cViT-M on the PACS and DomainNet datasets, following the same experimental settings as in Section~\ref{sec:eva:setup}. Computational cost is measured as the cumulative wall-clock time (in seconds) across all retraining.

\begin{figure}[t]
    \centering
    \includegraphics[width=\linewidth]{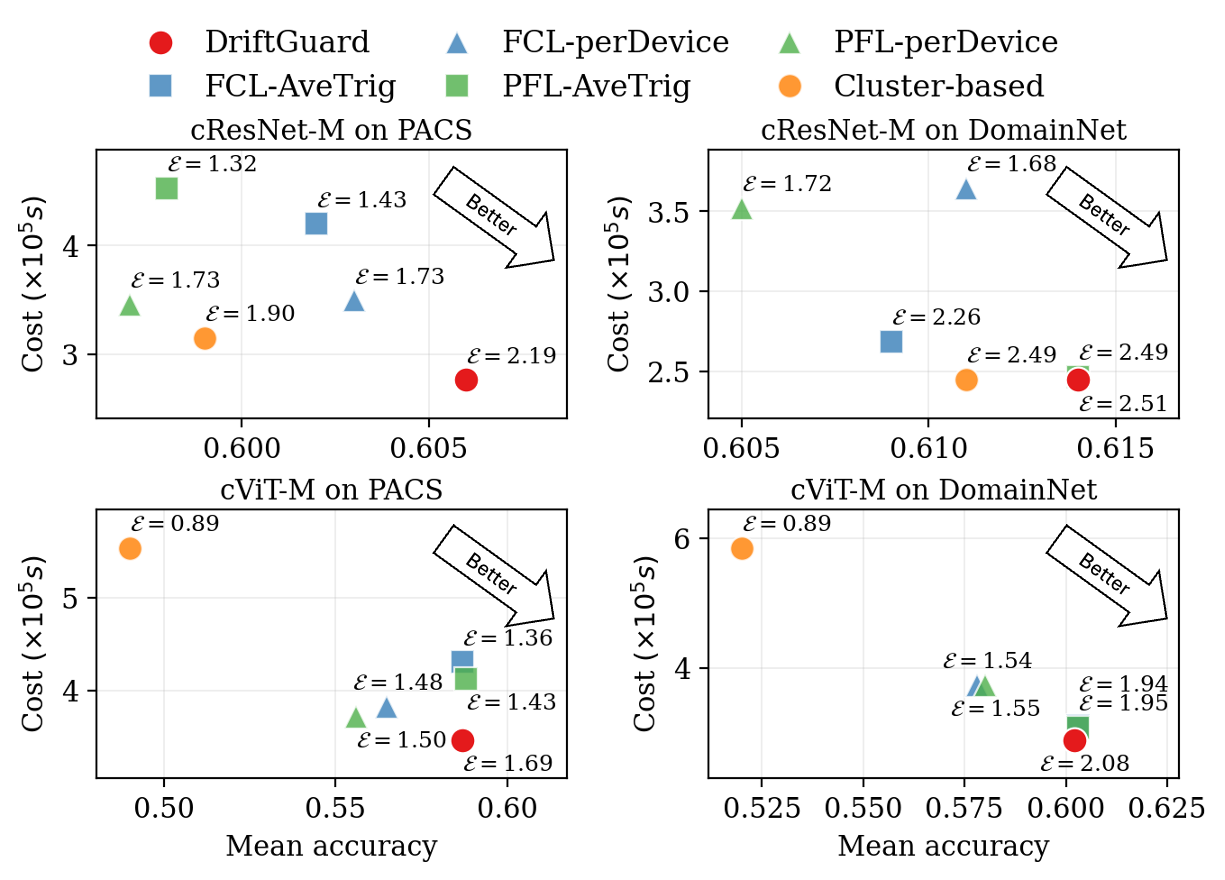}
    \caption{
        Mean accuracy versus total retraining time on the Raspberry Pi 4 prototype. Each point is annotated with efficiency $\mathcal{E}$, defined as mean accuracy per retraining time.
    }
    \label{fig:pi}
\end{figure}

Figure~\ref{fig:pi} presents the mean accuracy versus total retraining time for all methods on the Raspberry Pi prototype. Across all four configurations, \method achieves the highest or comparable accuracy while reducing retraining time by up to 20\% relative to the most cost-efficient baseline in each case. As a result, \method attains the highest accuracy per retraining time (\(\mathcal{E}\)) in all four configurations, up to 1.2$\times$ higher than the strongest baseline, consistent with the theoretical results in Section~\ref{sec:eva:result}. These results demonstrate that \method remains effective on real-world resource-constrained IoT hardware.

\subsection{Impact of hyperparameters in \method}
\label{sec:eva:ablation}
In this section, we study the impact of two key hyperparameters in \method: (i) the global retraining trigger threshold, and (ii) the distance threshold for group clustering. All experiments are conducted on DomainNet with cViT-M, and all other settings remain the same as those in Section~\ref{sec:eva:setup}.

\begin{figure}[t]
    \centering
    \includegraphics[width=\linewidth]{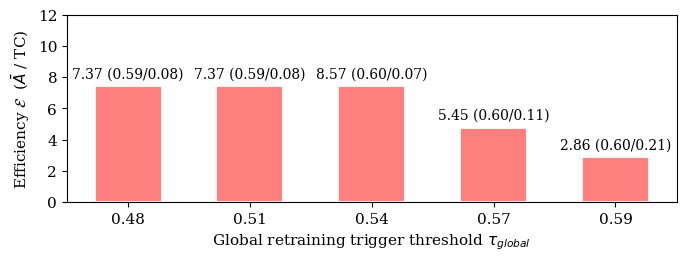}
    \caption{
    Impact of the global trigger threshold $\tau_{global}$ in \method. Each point is annotated with $\mathcal{E}$ ($\bar{A}$/TC), where $\mathcal{E}$ measures accuracy achieved per unit retraining cost.
    }
    \label{fig:ab1}
\end{figure}

\paragraph{Global retraining trigger threshold}
Global retraining in \method is triggered when the average accuracy across all devices falls below a threshold \(\tau_{global}\).
We vary \(\tau_{global}\) from 0.48 to 0.59 as shown in Figure~\ref{fig:ab1}.
As \(\tau_{global}\) increases, accuracy improves only marginally from 0.59 to 0.60, whereas retraining cost rises more substantially from 0.08 to 0.21. Specifically, the impact of \(\tau_{global}\) remains limited when \(\tau_{global} \leq 0.54\). However, once \(\tau_{global} > 0.57\), its effect becomes more pronounced: retraining cost increases more noticeably as global retraining is triggered more frequently, causing \method to behave similarly to \textit{FCL-AveTrig}.

\begin{figure}[t]
    \centering
    \includegraphics[width=\linewidth]{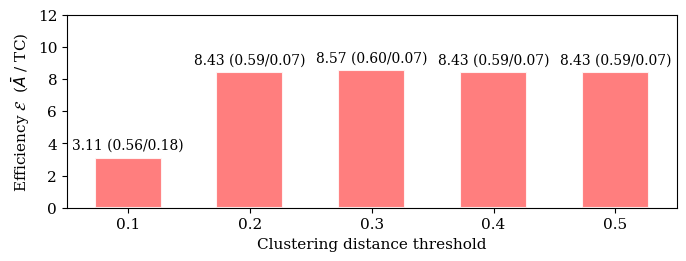}
    \caption{
        Impact of the clustering distance threshold. Each point is annotated with mean accuracy divided by retraining cost ($\bar{A}/TC$).
    }
    \label{fig:ab2}
\end{figure}

\paragraph{Clustering distance threshold}
The clustering distance threshold controls the granularity of grouping: smaller values produce a larger number of smaller groups, while larger values produce a smaller number of larger groups. We vary this threshold from 0.1 to 0.5, as shown in Figure~\ref{fig:ab2}.

Overall, the impact of this threshold is limited across most of the evaluated range. 
However, when the threshold is set to an extreme value (e.g., 0.1), it results in many small groups with only a few devices each. Such groups lack sufficient data for effective retraining, leading to more frequent retraining and higher retraining cost.
For thresholds between 0.2 and 0.5, \method remains stable, with accuracy ranging from 0.59 to 0.60 and efficiency ranging from 8.43 to 8.57.

%% file: sections/relatedwork.tex
\begin{table*}[t]
    \centering
    \small
    \setlength{\tabcolsep}{6pt}
    \renewcommand{\arraystretch}{1.15}
    \setlength{\tabcolsep}{3pt}
    \caption{Comparison of key features of existing FCL methods with \method under asynchronous drift.}
    \label{tab:rw_comparison}
    \newcommand{\cmark}{\ding{51}} 
    \newcommand{\xmark}{\ding{55}} 
    \begin{tabularx}{\linewidth}{Xccccc}
        \toprule
        \textbf{Features} &
        \textbf{
            \makecell{Classic FCL \\
            \cite{kirkpatrick2017ewc, lopez2017gem, mallya2018packnet}}
        } &
            \textbf{
                \makecell{Clustering / Multi-model FCL \\
                \cite{jothimurugesan2023feddrift, peng2025feddaa, Hao2025nazar}}
            } &
            \textbf{
                \makecell{Personalized FL \\
                \cite{yi2024moe, mei2024moe}}
            } &
            \textbf{\method} \\
        \midrule
        Adapting to asynchronous data drift 
            & \xmark & \cmark & \xmark & \cmark \\
        \makecell[l]{Decoupling global and personalized updates} 
            & \xmark & \xmark  & \cmark & \cmark \\
        Grouping devices with similar data distribution 
            & \xmark & \cmark & \xmark & \cmark \\
        Enabling lightweight adaptation
            & \xmark & \xmark  & \xmark & \cmark \\
    \bottomrule
    \end{tabularx}
\end{table*}

In this section, we review existing approaches related to \method, including classic \ac{FCL}, \ac{PFL}, and clustering-based approaches. Table~\ref{tab:rw_comparison} summarizes the comparison of key features across different approaches.

\subsection{Classic Federated Continual Learning}
Classic FCL extends continual learning techniques to federated settings. Existing methods includes regularization-based approaches~\cite{dong2022GLFC, dong2023LGA} that add penalty terms to prevent forgetting previously learned knowledge;
replay-based approaches~\cite{zhang2023target, wang2023FedPMR} that maintain memory buffers for rehearsal; and dynamic-architecture methods~\cite{yoon2021FedWeIT}, that expand model components to accommodate new patterns. 
However, these methods assume synchronous data drift where all devices experience similar drift patterns, and trigger global retraining involving all devices whenever drift is detected. This incurs high computational costs under asynchronous data drift.

\subsection{Personalized Federated Learning}
PFL decomposes models into shared and personalized components to address data heterogeneity~\cite{tan2022sv-pfl}.
Recent works, such as pFedMoE~\cite{yi2024moe} and FedMoE~\cite{mei2024moe}, adopt Mixture-of-Experts (MoE) architectures to further mitigate data heterogeneity.
However, existing PFL methods primarily target static non-IID settings and do not address data drift, where data distributions evolve dynamically over time.
\method builds upon the MoE-based decomposition in PFL, extending it to asynchronous data drift scenarios.

\subsection{Clustering-based Approach}
To address asynchronous data drift, recent FCL work proposes clustering-based strategies.
For example, FedDrift~\cite{jothimurugesan2023feddrift} formulates drift adaptation as a time-varying clustering problem and applies hierarchical clustering to dynamically group devices, maintaining separate models for each group.
FedDAA~\cite{peng2025feddaa} further introduces a dynamic clustered FL framework that distinguishes different types of drift and applies corresponding adaptation strategies.
While these approaches can separate devices with divergent data distributions, they treat groups as independent units and do not explicitly distinguish globally transferable knowledge from group-specific knowledge.
As a result, they may trigger repeated cluster-wise retraining even when multiple groups could benefit from shared updates.

%% file: sections/conclusion.tex

In this paper, we address asynchronous data drift in \ac{FL}, where devices experience distribution shifts at different times and rates. We propose \method, an \ac{FCL} framework that leverages an \ac{MoE}-inspired architecture to decouple shared and group-specific parameters, enabling global and group retraining.
Experiments show that \method achieves comparable or higher accuracy than the strongest baselines while reducing retraining cost by up to \costredT, resulting in up to \effimpt higher accuracy per cost ($\mathcal{E}$). On a real-world IoT prototype, it achieves up to 20\% lower retraining time while maintaining the highest accuracy. These results show that \method delivers a strong balance between accuracy and system overhead under asynchronous data drift, enabling efficient deployment on resource-constrained IoT devices.

%% file: sections/acknowledgment.tex
This research is supported by the UK Research and Innovation grant EP/Y028813/1.

%% file: sections/authorsbiography.tex
\newpage
\vskip -2.5\baselineskip plus -1fil
\begin{IEEEbiography}[{\includegraphics[width=1in,height=1.25in,clip,keepaspectratio]{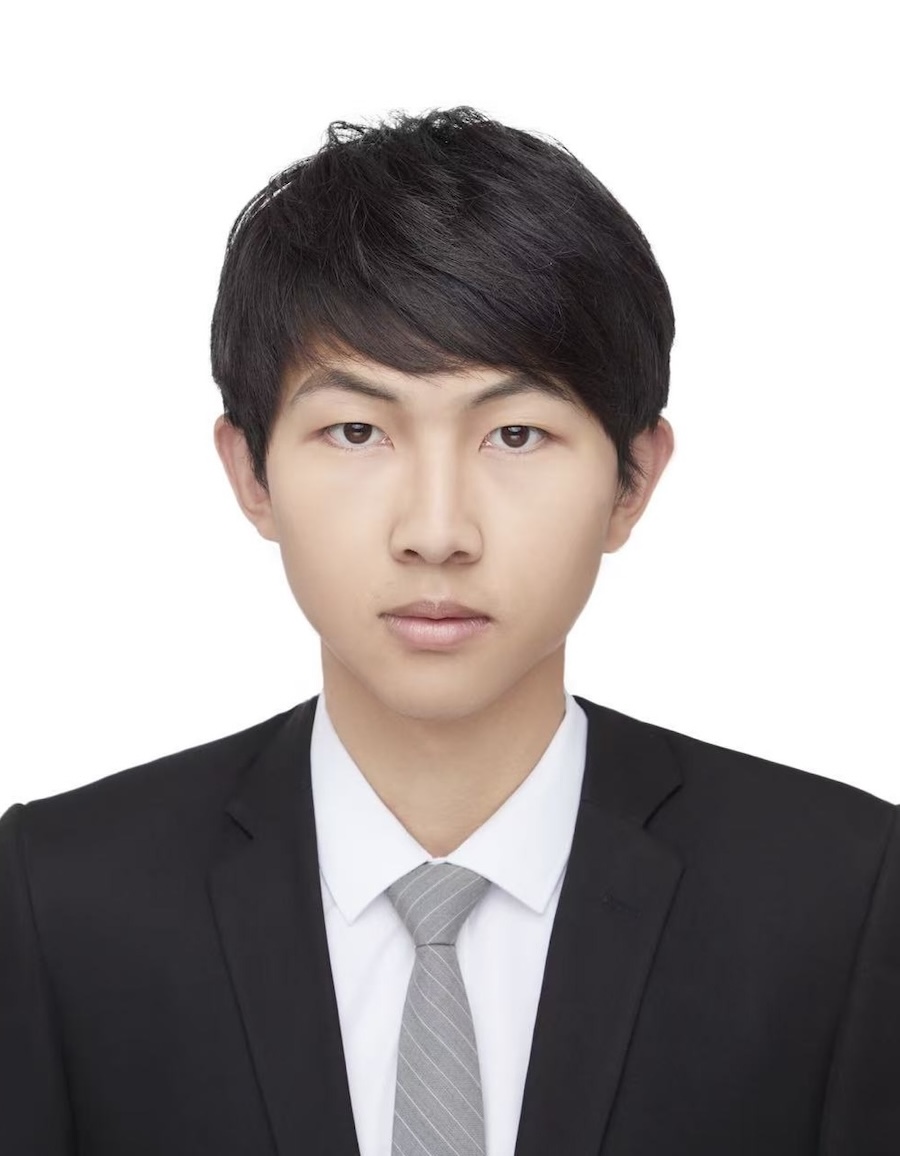}}]{Yizhou Han} received the MSc degree in Computer Science from the University of St Andrews, UK. His research interests include federated learning, AI inference, edge computing, and the Internet of Things (IoT). 
\end{IEEEbiography}
\vskip -2.5\baselineskip plus -1fil

\begin{IEEEbiography}[{\includegraphics[width=1in,height=1.25in,keepaspectratio]{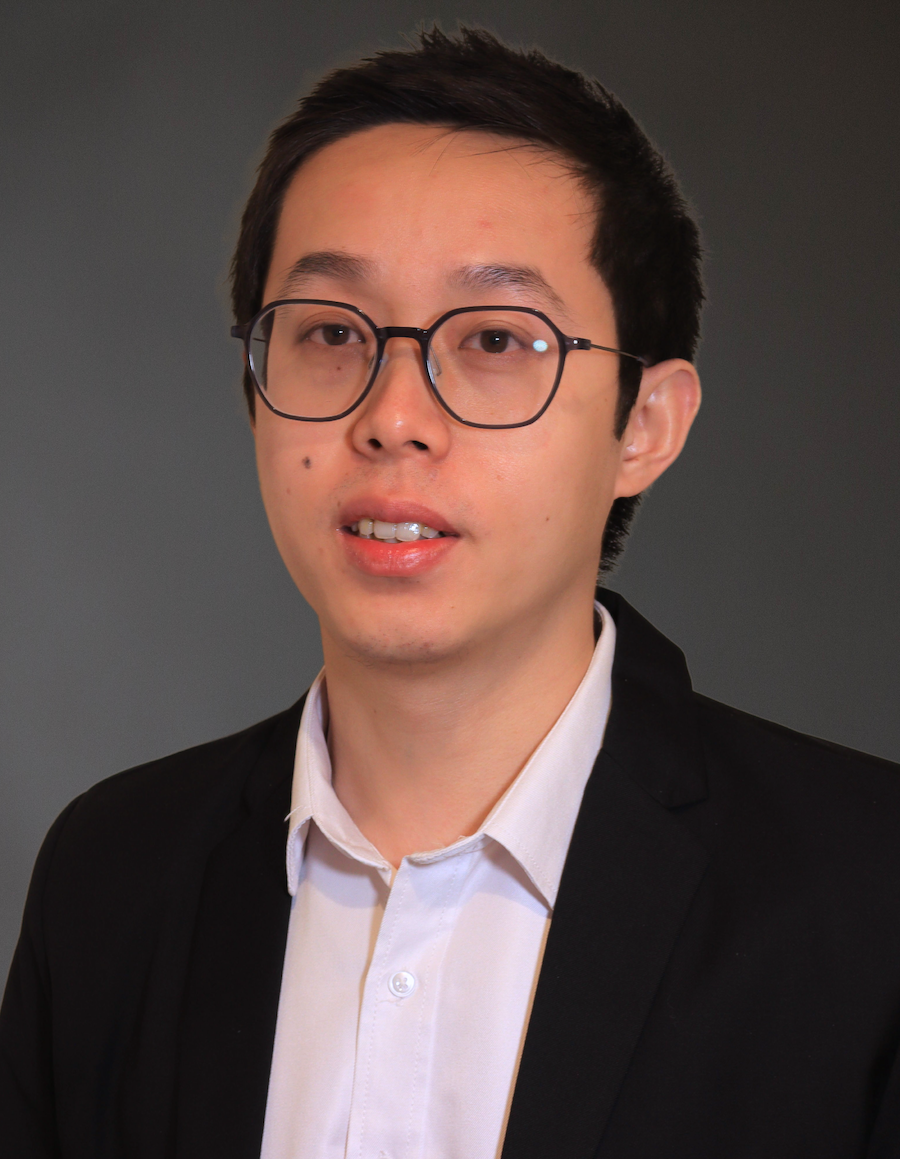}}]{Di Wu} is a Research Fellow at the University of St Andrews, UK. He received his PhD degree in Computer Science from the University of St Andrews, UK, in 2024. He is also a member of the Edge Computing Hub, St Andrews, and the National Edge AI Hub, UK. His primary research interests include federated learning, distributed machine learning, edge computing, and the Internet of Things (IoT).
\end{IEEEbiography}
\vskip -2.5\baselineskip plus -1fil

\begin{IEEEbiography}[{\includegraphics[width=1in,height=1.25in,keepaspectratio]{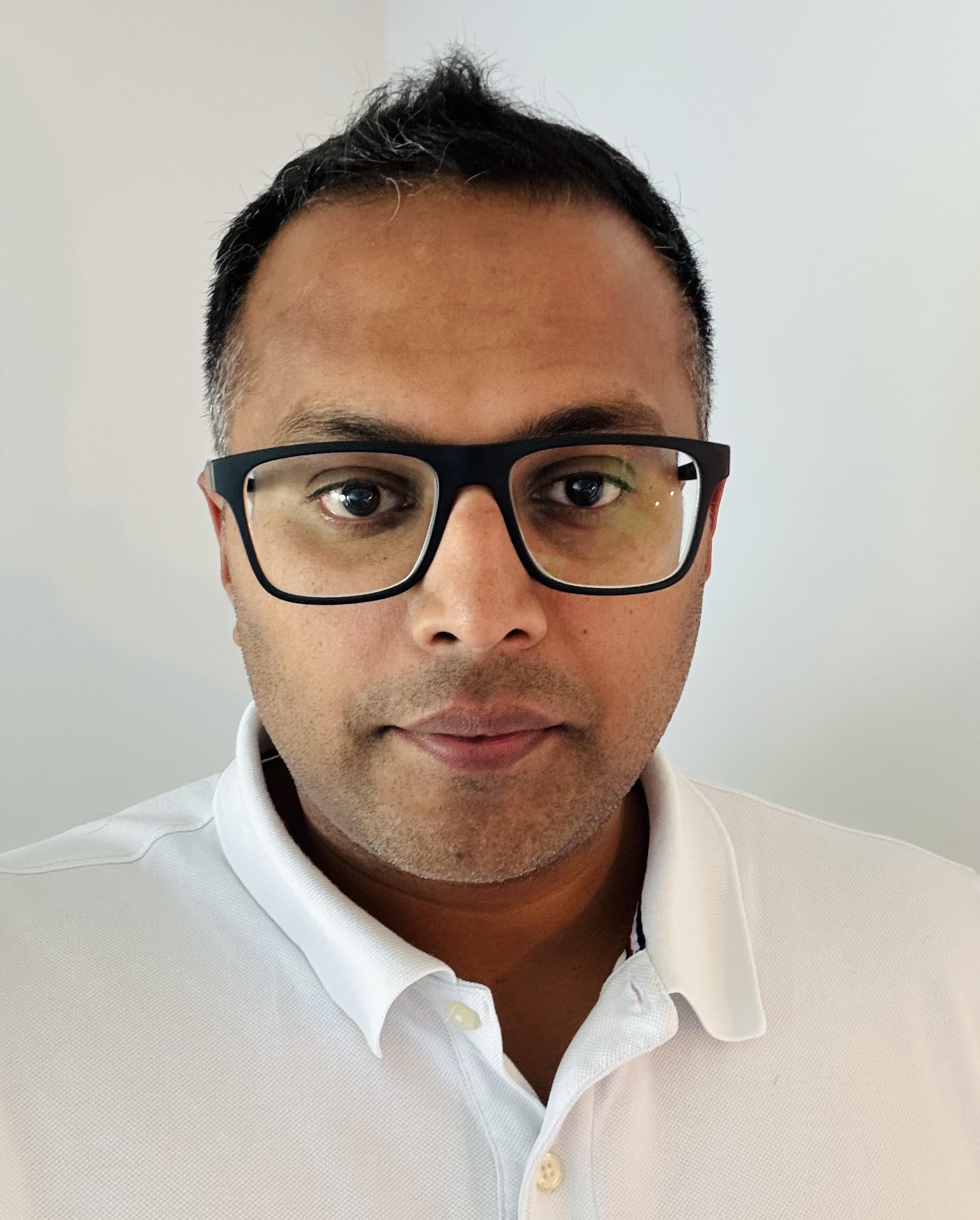}}]{Blesson Varghese} is a Professor in Computer Science at the University of St Andrews, UK, and the Director of the Edge Computing Hub. He leads research on the UK National Edge AI Hub. He is a previous Royal Society Short Industry Fellow. His interests include distributed systems that span the cloud-edge-device continuum and edge intelligence applications. More information is available from \url{www.blessonv.com}.
\end{IEEEbiography}
\vskip -2.5\baselineskip plus -1fil